\documentclass[letterpaper, 10 pt, conference]{ieeeconf}  

\IEEEoverridecommandlockouts                              

\usepackage{cite}
\usepackage{lipsum}
\usepackage{xcolor}
\usepackage{amsmath}
\usepackage{amssymb}
\usepackage{bm}
\usepackage{graphicx}
\usepackage[normalem]{ulem}
\usepackage{balance}

\newcommand{\eg}{\emph{e.g.},}
\newcommand{\ie}{\emph{i.e.},}

\def\tabref#1{Table~\ref{#1}}

\title{\LARGE \bf
Demonstrating Autonomous 3D Path Planning on \\ a Novel Scalable UGV-UAV Morphing Robot
}

\author{Eric Sihite$^{1}$, Filip Slezak$^{1}$, Ioannis Mandralis$^{1}$, Adarsh Salagame$^{2}$, Milad Ramezani$^{3}$, \\ Arash Kalantari$^{4}$, Alireza Ramezani$^{2*}$, and Morteza Gharib$^{1}$
\thanks{$^{1}$Authors are with the Department of Aerospace Engineering, California Institute of Technology, Pasadena, USA. Emails: 
        {\tt\small esihite, fslezak, imandralis, mgharib@caltech.edu}}%
\thanks{$^{2}$Authors are with the Silicon Synapse Labs, Department of Electrical and Computer Engineering, Northeastern University, Boston, USA. Emails: 
        {\tt\small salagame.a, a.ramezani@northeastern.edu}}%
\thanks{$^{3}$Author is with the Robotics and Autonomous Systems, DATA61, CSIRO, Brisbane, QLD 4069, Australia. Email: 
        {\tt\small milad.ramezani@data61.csiro.au}}%
\thanks{$^{4}$Author is with the Jet Propulsion Laboratory, Pasadena, USA. Email: 
        {\tt\small arash.kalantari@jpl.nasa.gov}}%
\thanks{$^{*}$Corresponding author.}%
}

\begin{document}

\maketitle
\thispagestyle{empty}
\pagestyle{empty}

\begin{abstract}

Some animals exhibit multi-modal locomotion capability to traverse a wide range of terrains and environments, such as amphibians that can swim and walk or birds that can fly and walk. This capability is extremely beneficial for expanding the animal's habitat range and they can choose the most energy efficient mode of locomotion in a given environment. The robotic biomimicry of this multi-modal locomotion capability can be very challenging but offer the same advantages. However, the expanded range of locomotion also increases the complexity of performing localization and path planning. In this work, we present our morphing multi-modal robot, which is capable of ground and aerial locomotion, and the implementation of readily available SLAM and path planning solutions to navigate a complex indoor environment.

\end{abstract}

\section{Introduction}




Robotic multi-modal locomotion can be a significant ordeal. The prohibitive design restrictions include a tight power budget, limited payload, complex multi-modal actuation, excessive number of active and passive joints involved in each mode, sophisticated control, autonomy, and environment-specific models (since different environments are involved), to name a few, which have alienated these concepts. That said, the number of designs surrounding ground-aerial locomotion is not small. The robotic community has endorsed the importance of these systems and tirelessly introduces interesting and novel concepts \cite{araki_multi-robot_2017, Flying_star, peterson2011experimental, tagliabue2020shapeshifter, sihite2021unilateral, liang2021rough, dangol2021hzd, dangol2021control, sihite2021optimization, ramezani2021generative}. 
However, these systems are too small to carry heavy perception pieces of equipment. The most notable examples are legged-wheeled systems \cite{schwarz2016hybrid, suzumura2013real, thomson2012kinematic, grand2004stability, bjelonic2019keep} which can carry large equipment for autonomy but their modes of mobility (legged and wheeled) do not lead to significantly different mobility capabilities in terms of environment traversability.

\begin{figure}[t]
\vspace{0.08in}
    \centering
    \includegraphics[width=\linewidth]{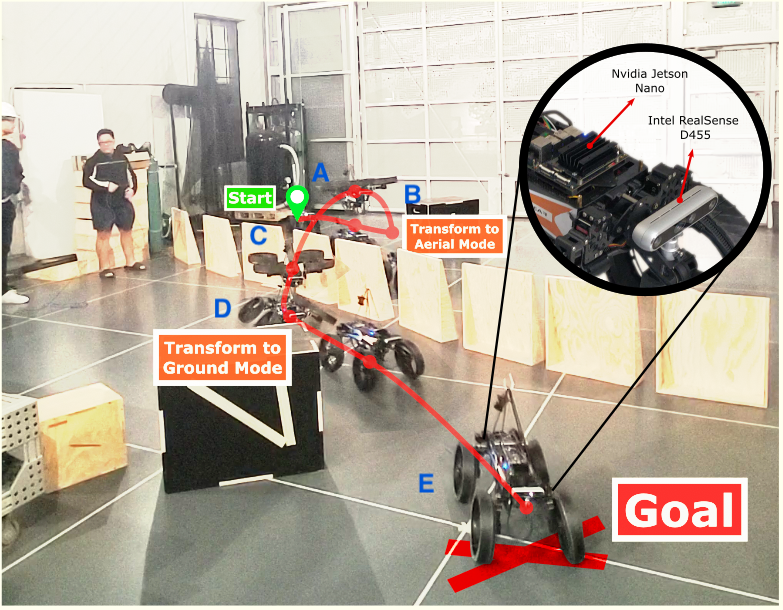}
    \caption{Composite image showing the path taken by the robot as it autonomously navigated the walled environment. The robot flew over the wall in UAS mode, transformed into UGV mode, then autonomously drove to the final waypoint. ABCDE letters correspond to specific actions time stamped in Fig \ref{fig:odometry}.}
    \label{fig:composite}
\vspace{-0.5cm}
\end{figure}

\begin{figure}[t]
\vspace{0.08in}
    \centering
    \includegraphics[width=0.85\linewidth]{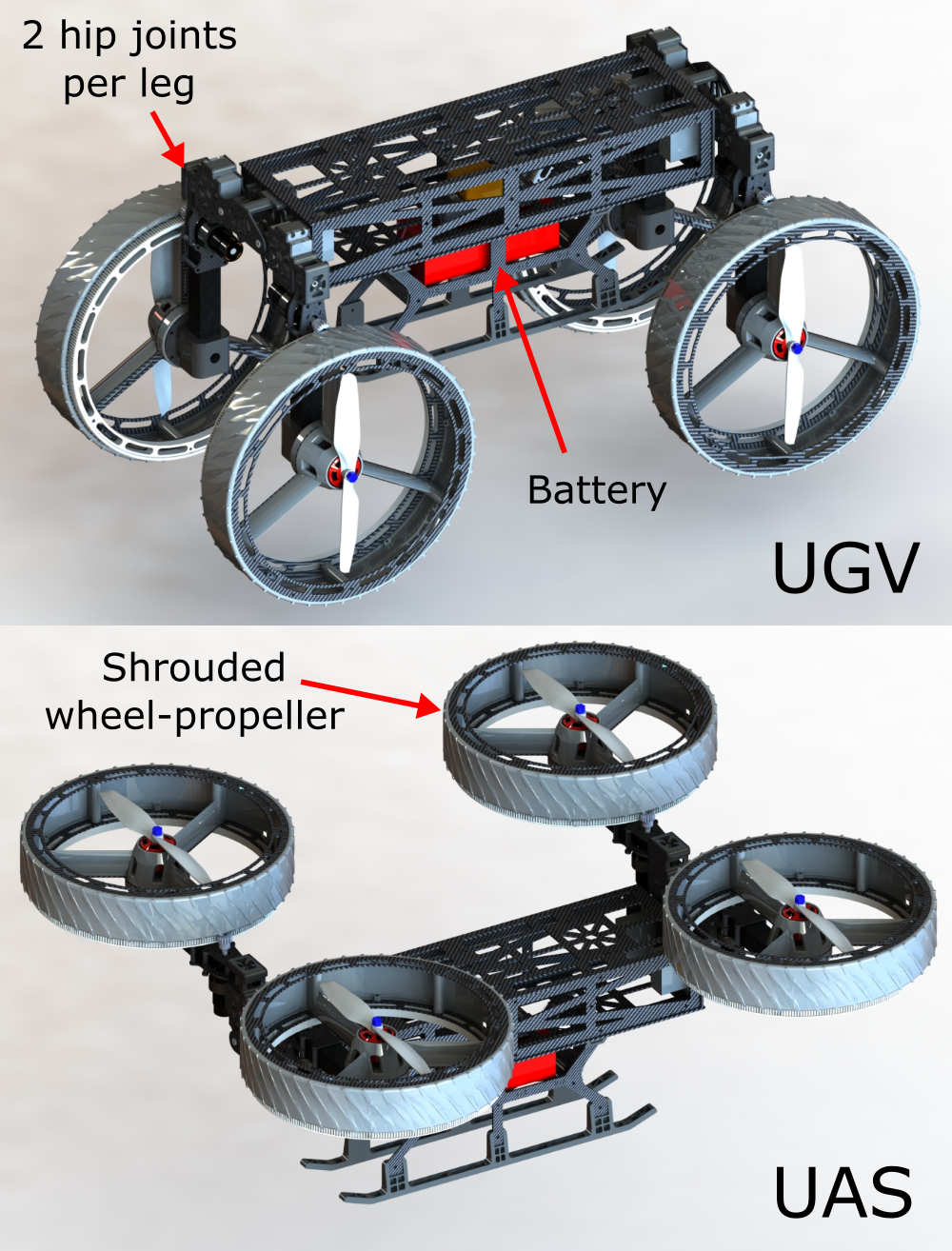}
    \caption{Illustrations of our multi-modal morphing robot that is capable of transforming between the unmanned ground vehicle (UGV) and unmanned aerial system (UAS) configurations.}
    \label{fig:robot_overview}
\vspace{-0.5cm}
\end{figure}

\begin{figure}[t]
\vspace{0.08in}
    \centering
    \includegraphics[width=0.9\linewidth]{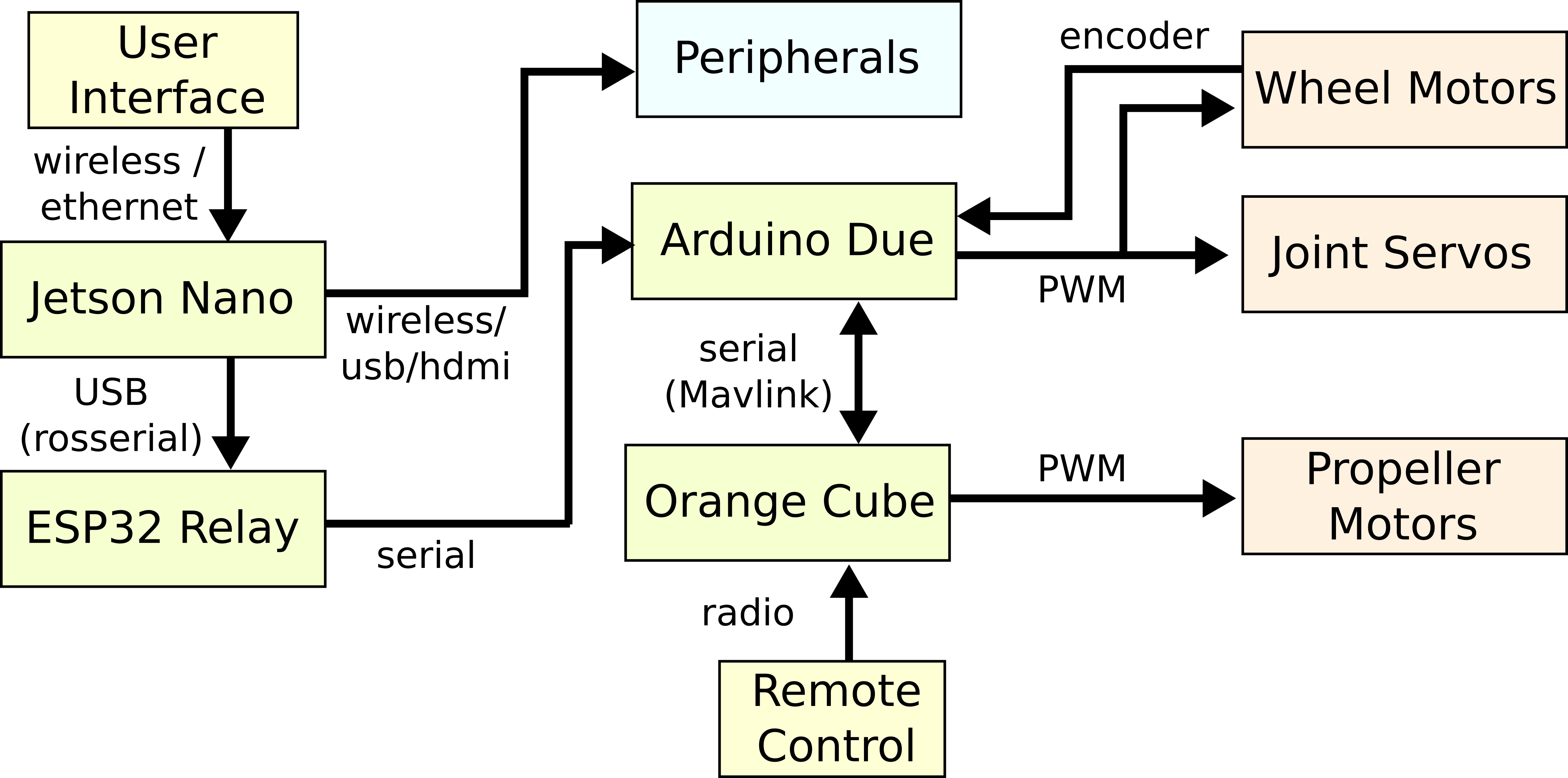}
    \caption{The robot's system architecture for control, sensing, and SLAM.}
    \label{fig:system_architecture}
\vspace{-0.5cm}
\end{figure}

This paper provides an overview of our robot, called the M4 \cite{sihite2023multi, mandralis2023minimum}, which possesses several modes of mobility and can carry large payloads (scalable design). M4's scalability positions it in a unique situation even when only two of its mobility forms, ground and aerial, are considered. M4 can be employed in many applications such as Search And Rescue (SAR), planetary exploration, last mile package delivery, and emergency payload transport operations. In these operations, mobile robots face different environments. For instance, in SAR operations and in the aftermath of unique incidents such as flooding, one event may accompany another disaster. A hurricane may produce flooding and wind damage, or a landslide may dam a river and create a flood. Therefore, the more locomotion plasticity a robot possesses, the higher the chance of success. 


Perception, localization, path-planning, and high-level decision-making have been extensively developed in robotics, and we recognize this fact. However, these advances have remained limited to ground-aerial vehicles that cannot carry large payloads. If we limit ourselves to two modes of operations simultaneously, the ground-aerial robots introduced so far are systems that are limited to small payloads. These systems have remained small due to the conflicting requirements of ground-aerial locomotion,~\ie~they are not scalable. They cannot carry extra payloads such as perception sensors like cameras, lidar sensors, and powerful enough computers needed for self-contained, fully autonomous operations in a similar fashion that legged robots -- which have scalable designs thanks to their operational nature based on contact forces -- perform autonomous tasks today. It can be seen that even when the modes of operation are limited to ground and aerial locomotion, the notions of autonomy are less explored.


This paper briefly overviews joint collaborative work between Caltech, Northeastern University, and Jet Propulsion Lab (JPL) in designing a platform called M4 with enhanced locomotion plasticity. In this paper, only ground-aerial locomotion is employed for autonomous multi-modal mobility. Extensive technical content about mechanism design and hardware aspects is skipped as the main goal is to limit the current draft to autonomy implementations of M4 in its ground-aerial modes only skipping other capabilities. The authors recognize the results presented here do not contribute to robot autonomy from a technical standpoint or introduce new algorithms. However, this work is distinguished in the following ways; 
{
The ability to realize ground and aerial mobility in a scalable fashion (\ie~the payload can be large) has never been achieved before, and ground-aerial systems introduced so far are limited to small systems with very restricted sensing and computing capabilities. The opportunity offered by a scalable multi-modal robot that can carry large electronics despite the conflicting requirements dictated by ground and aerial locomotion unlocks unexplored paths, particularly in robot perception, localization, and decision-making at a totally different level. For instance, often mobile robots are limited to fixed constraints in terms of their perception ranges. Following successful DARPA sub-T stories, adding robot hitch-hikers such as quadcopters to ground robots has been widely utilized. The multi-modal robot recruited in this paper combines these opportunities in a single, scalable platform, avoiding all other challenges associated with robot hitch-hiker concepts (\eg~data transfer, homing, docking, etc.) yet offering sub-T-like capabilities. Future autonomy and path planning for M4 will focus not just on the binary choice of aerial versus ground locomotion but on planning paths that utilize the full capabilities of M4. This paper, however, focuses on the initial steps towards this eventual goal by demonstrating autonomous ground-aerial locomotion.}



\section{Mechanical Design and System Overview}

\begin{table}[t]
\vspace{0.08in}
\caption{M4 Hardware Configuration}
\centering
\begin{tabular}{lccr}
\hline
\multicolumn{4}{c}{Configuration Size}                                           \\ \hline
\multicolumn{1}{l}{UGV}                    & & & 0.70$\times$0.35$\times$0.35 $m^3$                   \\
\multicolumn{1}{l}{UAS}                    & & & 0.70$\times$0.70$\times$0.30 $m^3$                   \\
\multicolumn{1}{l}{MIP}                    & & & 1.00$\times$0.50$\times$0.30 $m^3$                   \\ \hline
\multicolumn{1}{l}{Thrust/Weight}          & & & 9kg/6kg                           \\ \hline
\multicolumn{4}{c}{Actuators}                                                    \\ \hline
\multicolumn{1}{l}{Wheels}                 & & & 4 motors                          \\
\multicolumn{1}{l}{Rotors}                 & & & 4 motors                          \\
\multicolumn{1}{l}{Joints}                 & & & 8 servos                          \\ \hline
\multicolumn{1}{l}{Power}                  & & & 6S 4000mAh LiPo                   \\ \hline
\multicolumn{4}{c}{Compute Units}                                                    \\ \hline
\multicolumn{1}{l}{Perception \& Navigation}      & & & Nvidia Jetson Nano                \\
\multicolumn{1}{l}{Wheels \& Joints Control}&& & Arduino Due                       \\
\multicolumn{1}{l}{Rotors Control}         & & & Cube Orange                       \\
\multicolumn{1}{l}{Communication Unit}           & & & ESP-WROOM-32 \\ \hline
\end{tabular}
\label{tab:hardware_config}
\vspace{-0.5cm}
\end{table}

Our transforming robot, shown in Fig.~\ref{fig:robot_overview}, can switch its modes of mobility between the unmanned ground vehicle (UGV), unmanned aerial system (UAS), mobile inverted pendulum (MIP), quadrupedal, thruster-assisted MIP, legged locomotion, and manipulation. The robot possesses an articulated body with four legs, each leg has a total of two hip degrees of freedom (DOF) and a shrouded propeller that act as a wheel and a thruster simultaneously. Eight independent joint actuators translate the legs forward, backward, and sideways, and the shrouded propellers are attached to the leg ends. Additionally, it is equipped with an Intel RealSense D455 stereo depth (RGB-D) camera for perception, localization, and mapping.

As summarised in~\tabref{tab:hardware_config}, the robot weighs approximately 6.0 kg with all components, which include the onboard computers for low-level control and data collection, sensors (encoders, inertial measurement unit, stereo cameras), communication devices for teleoperation, and power electronic components. Most of the robot's weight stems from the high-power components consisting of 4 wheel motors, 4 propeller motors, 8 joint servos, 8 motor drivers, and a 6S 4000mAh battery. When in UGV mode, the robot measures 0.7 m in length and 0.35 m in both width and height. When in the MIP mode and dynamically balancing on its two wheels, it is 1.0 m tall, which permits reaching a better vantage point for data collection using its exteroceptive sensors. When in UAS configuration, the robot is 0.3 m tall, and the propellers' center points can reach a maximum distance of 0.45 m far apart from each other. Each propeller-motor combination can generate a maximum thrust force of approximately 2.2 kg-force, therefore reaching roughly 9 kg thrust force in total for an approximately 1.5 thrust-to-weight ratio. Its legs are 0.3 m long, including its 0.25 m in diameter wheels, which allows for traversing bumpy terrain.


Figure \ref{fig:system_architecture} shows the system architecture and data pipelines of the robot. The robot utilizes a Jetson Nano (Quad-core ARM A57 CPU and 128-core Maxwell GPU) computer, and several microcontrollers in the system: Arduino Due (32-bit Atmel SAM3X8E ARM Cortex-M3), Orange Cube flight controller (32-bit STM32H753 ARM Cortex-M7), and ESP-WROOM-32 (Xtensa 32-bit LX7 dual-core processor). Each microcontroller performs specific tasks and communicates with the others through serial. The Arduino Due handles the communication between all components and drives all actuators except the thruster motors. The Orange Cube controls the robot's aerial mobility using the open-source ArduPilot framework and communicates with the Arduino through MAVLink messaging protocols. The Jetson Nano is a powerful and small computer used to process the stereo depth camera data and SLAM, which then are relayed to the flight controller. Finally, the ESP32 acts as a simple relay between the Jetson Nano and Arduino Due due to incompatibilities with rosserial that sometimes triggered the system to reset, which can be dangerous if happened mid-flight.

\section{SLAM and Path Planning}


In this section, we describe two key aspects of mobile robotics: sensing and modeling the environment and then planning action based on that information. For the M4 we leverage a V-SLAM approach for state estimation and a multi-modal probabilistic roadmap (MM-PRM) for a multi-modal approach to localization and navigation.

\subsection{Simultaneous Localization and Mapping}

Simultaneous Localization and Mapping (SLAM) and path planning are crucial concepts in autonomy. SLAM refers to estimating the robot's pose (in our case $\mathbf{x}\in\mathbb{R}^6$) and mapping the environment $\mathcal{M}$ simultaneously, typically by measuring the rotation of wheels or other locomotion mechanisms. Path planning, on the other hand, involves finding the optimal path for a robot (relying on the SLAM output) to reach a desired destination or complete a task.

If SLAM is inaccurate, path planning algorithms may generate suboptimal paths or cause the robot to deviate from its intended path, which can lead to errors or even collisions. In this work, to achieve accurate SLAM measurements, we use Real-Time Appearance-Based Mapping (RTAB-Map)~\cite{labbe2019rtab} based on the integration of RGBD and inertial measurements. In our system's SLAM module, ORB/GFTT visual features~\cite{orb, gftt} are extracted, and a bag of words~\cite{bow} is used for loop closure detection.

Once accurate SLAM measurements are obtained, path planning algorithms can be used to generate the most efficient path for a robot to reach its goal. These algorithms consider factors such as obstacle avoidance, terrain conditions, and energy consumption.
One common approach to path planning is the use of algorithms such as A* or Dijkstra's algorithm~{\cite{A_star, dijkstra1959note, comboplanner}}, which generates a graph of the environment and calculates the shortest path between the robot's current position and its goal. These algorithms can be further optimized by incorporating heuristics or other techniques~{\cite{visgraph, Astar_heuristics, Astar_time}}.

\begin{figure*}[t]
\vspace{0.08in}
    \centering
    \includegraphics[width=0.8\linewidth]{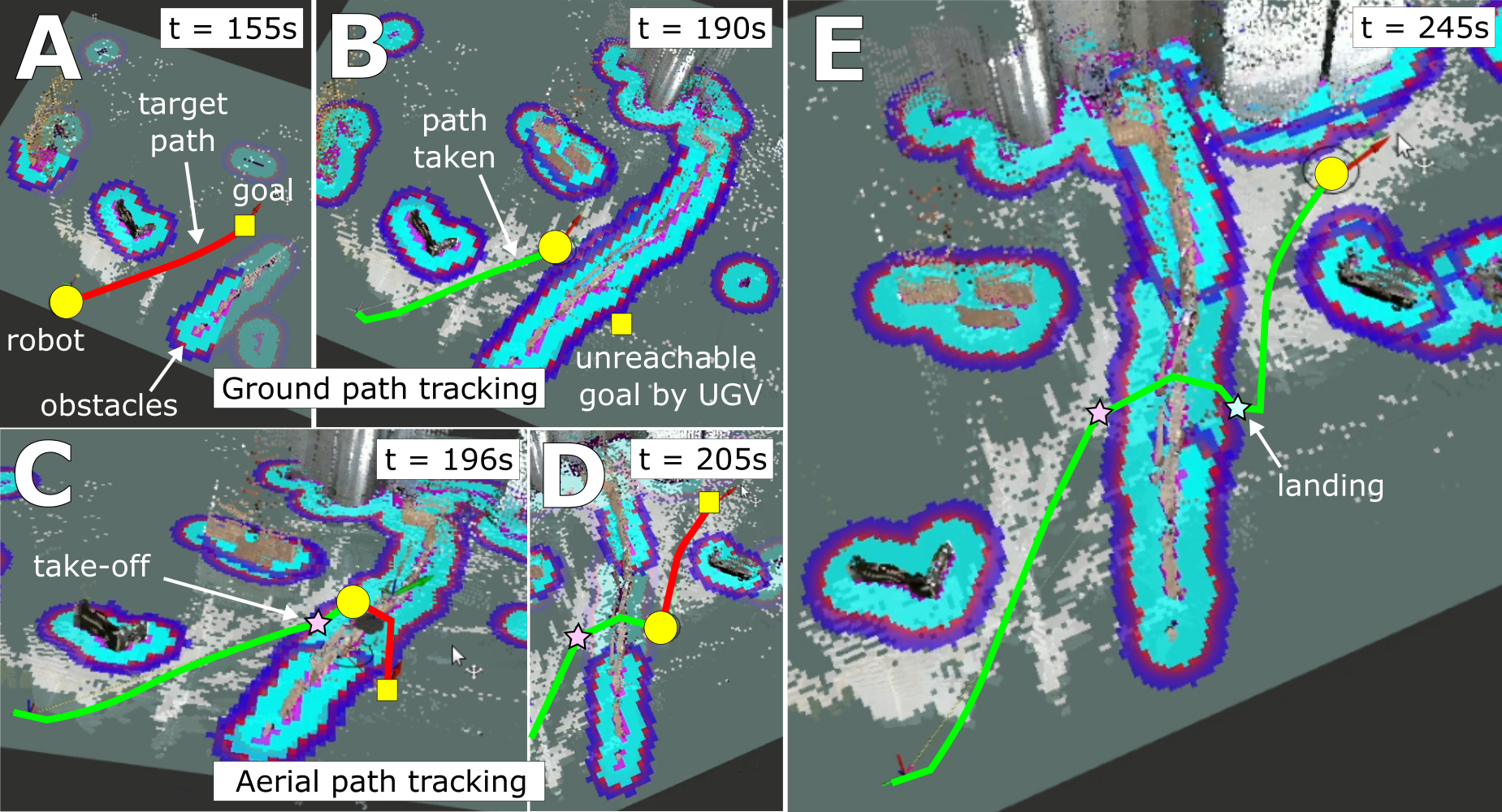}
    \caption{Visualization of the SLAM algorithm as the robot explored and followed the desired waypoints. (A) and (B) show the online ground waypoint generation using A$^*$ algorithm and the robot's ground trajectory tracking. In (B), a waypoint that can't be solved by the A$^*$ was given to the robot. It must transform into UAS mode and fly over the obstacles to reach the waypoint, as shown in (C) and (D). Then once the robot has landed, it transformed back into UGV mode and navigated itself towards the final waypoint, as shown in (D) and (E).}
    \label{fig:odometry}
\vspace{-0.5cm}
\end{figure*}

{
\subsection{High-level Decision Making and Path Planning}

The objective of the path planning strategy is to minimize the total energy consumed by the robot by prioritizing ground over aerial locomotion. To achieve this, the path planning algorithm set up the environment by discretizing it into nodes where each node is associated with a locomotion mode (ground or aerial). The path planning module in our system represents the environment as a graph $\mathcal{G}=(\mathcal{V}, \mathcal{E})$, where $\mathcal{V}\in\{\mathbf{x}\}^n_{i=1}$ are $n$ possible robot poses within the graph. 
~$\mathcal{E}$ denotes a set of edges between nodes. 



Following the work done in \cite{sihite2022efficient, MM_PRM}, the 3D environment is discretized into a set of nodes and edges with the 3D multi-modal probabilistic roadmap (MM-PRM) which takes into account the multi-modal nature of the robot's locomotion. The classical PRM algorithm builds a graph in the defined space by generating a certain number of nodes, where each node is created with a random position. When a node is created, it will search for the nearest nodes already present in the graph and then connect to them to form edges while checking that it does not cross any obstacles. This classical method can be extended into the 3D environment by creating ground and aerial nodes separately, then connecting the nodes with edges to form a single multi-modal environment. This version of the PRM algorithm requires the definition of 3 parameters: the number of ground surface nodes $N_w$, the number of nodes describing flyable space $N_f$, and the maximum distance between neighboring nodes $R$.

New ground nodes $\mathbf x_{new}$ are randomly assigned according to the following constraint:
\begin{equation}
\mathbf x_{new} \in \{(x, y, z): z=z_{GND})\},
\end{equation}
\noindent where $x$, $y$, and $z$ represents the node position in Euclidean space, and $z_{GND}$ is the ground elevation. Similarly, new nodes in the flyable task space are obtained as follows:
\begin{equation}
\mathbf x_{new} \in \{ (x, y, z): z > 0, z \ne z_{GND} \}.
\end{equation}
The edges ($E$) are created by searching for the neighboring nodes, which are found using the following condition:
\begin{equation}
    \mathbf x_{Nearest} = \{\mathbf x \in \mathcal{N} : \| \mathbf x_{new} - \mathbf x \| \leq R\},
\end{equation}
\noindent where $\mathcal{N}$ is the set of nodes already created, $R$ is the maximum radius distance, and $\| . \|$ is the Euclidean norm.

\subsubsection{Calculation of Locomotion Cost}
\label{sec:cost_calculation}


To calculate the locomotion cost including legged and aerial, it is necessary to not only determine the costs associated with each modes but also the cost corresponding to the transition from one mode to another. As such, the cost of transport on a ground edge $C_g$ is calculated using the motor power consumption $P_m$, which is integrated over the time of wheeled locomotion. The total joint power consumption is computed based on the torque and the angular velocity of each joint. The time of legged locomotion is calculated based on the distance $d$ between the two nodes. As a result, $C_g$ is given by:
\begin{equation}
    \textstyle
    C_g =  \int_0^{t_d} P_m(\tau)d\tau.
\end{equation}
%
The energetic cost on a flying edge $C_f$ is computed using the power consumption $P_f$ in hovering, the robot forward velocity $v_f$ in flying mode, and the altitude $z$ of the two nodes. Hence, $C_f$ is given by:
\begin{equation}
    C_f = P_f \, (d / v_f) + mg(z_2 - z_1),
    \label{eq_Cf}
\end{equation}
\noindent where $z_1$ and $z_2$ are respectively the altitudes of the nodes 1 and 2, $m$ is the mass of the robot and $g$ is the gravitational acceleration constant. Last, the transition cost $C_t$ between the two modes is determined based on the power consumption of the joints during the morphing process $P_t$. Then, $P_t$ is integrated over the time of transition $t_t$ which yields:
\begin{equation}
\textstyle
    C_t = \int_0^{t_t} P_s(\tau)d\tau .
\end{equation}
\noindent These three energetic costs are employed to determine the optimal path in the edge space generated by MM-PRM algorithm using the A$^\star$ algorithm.

\subsubsection{Optimal Path Using 3D A* Algorithm}

\begin{figure*}[t]
\vspace{0.08in}
    \centering
    \includegraphics[width=0.45\linewidth]{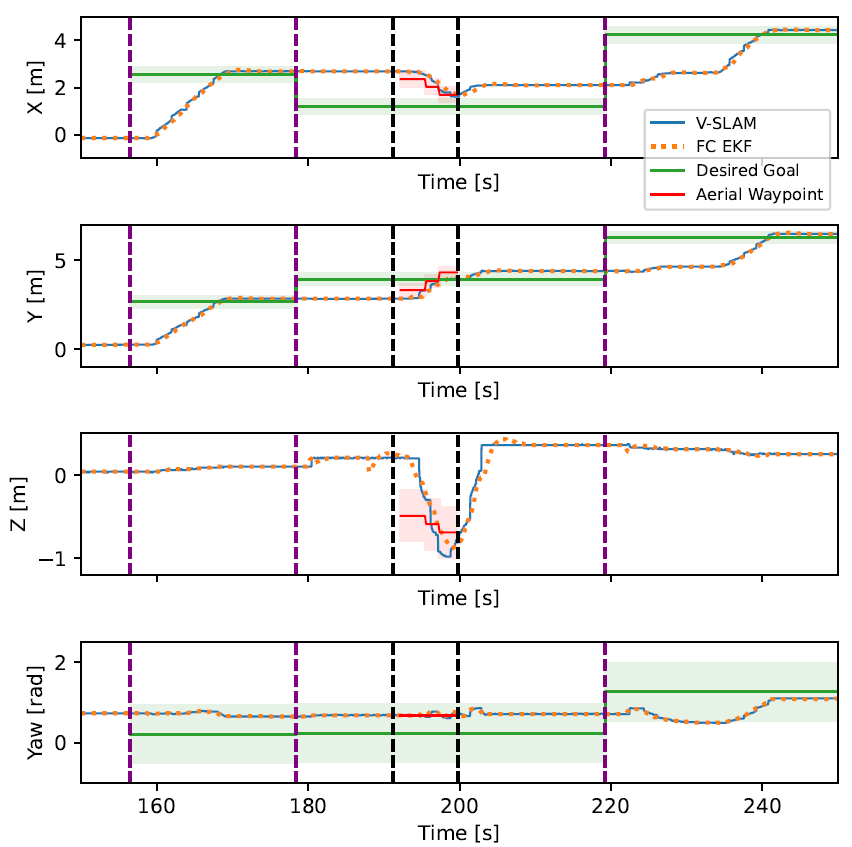}
    \hfill
    \includegraphics[width=0.45\linewidth]{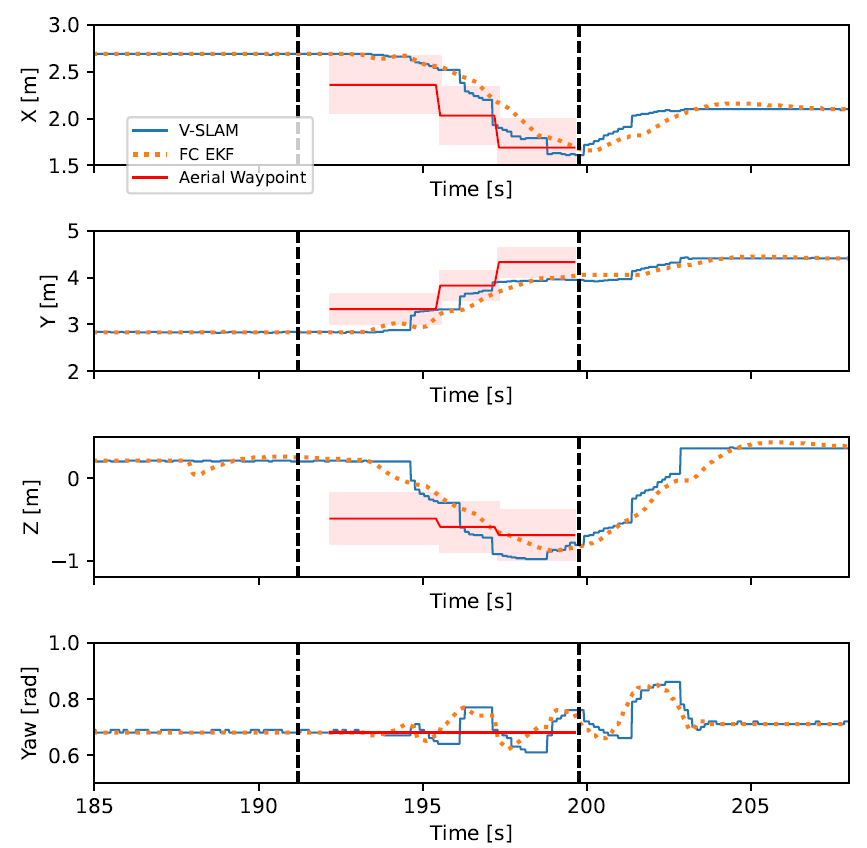}
    \caption{Plots of the robot's pose as estimated by the stereo depth camera Visual SLAM (V-SLAM) localization and the target waypoints. The left plot shows the states throughout the entire experiment while the right plot shows the states during aerial mobility. The EKF estimate from the flight controller (FC) is also provided to show the delay between stereo depth camera's and the FC's internal pose estimates. The shaded region illustrates the tracking margin of error in the controller.}
    \label{fig:odometry_plot}
\vspace{-0.5cm}
\end{figure*}

To find the optimal path in the graph, the A$^\star$ path search algorithm \cite{A_star, dijkstra1959note} is used with a heuristic function adapted to the robot's multi-modal abilities.
Each time the algorithm explores $n$-th node, it calculates the minimum cost for the A$^*$ objective function $g(\mathbf{x}) + h(\mathbf{x},\mathbf{x}_g)$.
The heuristic cost $h(\mathbf{x},\mathbf{x}_g)$ is calculated by summing the following two costs. First, the cost of walking on flat ground to the goal in a straight line is calculated. Second, the cost of flying vertically along the z-axis to the goal is obtained. Since the cost of walking is much lower than flying, this is the most optimal way to move between two points if there is no obstacle or impassable terrain between the current and target positions. The following cost for $g(\mathbf{x})$ is defined:
\begin{equation}
    \textstyle
    g(\mathbf{x}) = \sum_{i = 0}^{E_w} C_{w,i} + \sum_{j = 0}^{E_f} C_{f,j} + N_t\, C_{t}
    \label{eq:g(n)}
\end{equation}
\noindent where $E_w$ and $E_f$ are the number of walking and flying edges traveled by the robot, respectively. Furthermore, $C_{w,i}$ is the cost on the walking edge $i$, $C_{f,j}$ is the cost on the flying edge $j$, and $N_t$ is the number of mode transitions made by the robot (ground to aerial, or vice versa).
}

For collision avoidance, we benefit from the Dynamic Window Approach (DWA)~\cite{fox1997dynamic}, which casts several paths in front of the robot (based on the robot kinematic configuration) and scores them based on some criteria,~\eg~obstacle proximity, goal proximity. The path with the maximum score is the optimum path toward the desired goal.

\section{Experimental Results and Discussions}


A set of experiments were conducted inside the California Institute of Technology's CAST Arena. An obstacle course was set up for the robot to perform localization, path planning, and trajectory tracking using the robot's ground and aerial multi-modal locomotion. The goal of this experiment is to show the effectiveness of the navigation stack, and the versatility of our transforming robot where it can fly over difficult or unpassable terrains. Therefore, a "walled" area was set up where the robot must fly over the wall to reach the target position, as illustrated in Fig.~\ref{fig:composite}. 

For this autonomous navigation task, an operator will provide a target position to the robot through RViz and the robot will then figure out a way to reach that position. The system relies on a single stereo depth camera (Intel RealSense D455, which has an onboard IMU) fixed at the front end of the robot. The information is then processed on the Jetson Nano running ROS software. The RTAB-Map SLAM algorithm relies on VIO to reconstruct a point cloud representation of the environment and provide pose estimates of the robot. This point cloud is then filtered by object height and projected onto the driving plane and obstacles are inflated to provide a collision safety margin for which the robot is assimilated to a sphere in space. The resulting map is then used by MoveBase A* global planner and DWA (dynamic window approach) local planner for 2.5D ground navigation. When a path to the desired target exists, motor commands are transferred to the main ground controller for execution. Then, the map and pose estimates are continuously updated and a path is recomputed live to account for new observations or displacement errors. If a path cannot be found, the robot assumes that an aerial path exists, based on the operator's knowledge of the environment, and will trigger a flight to its target. As soon as that signal is received, the robot will morph into UAS mode and receive the target position to generate guided mode waypoints for the ArduPilot flight controller. Then a state machine will communicate with ArduPilot over MAVLink to fly the robots to the desired waypoints with a straight line and fixed height trajectory to the target. After landing, the robot awaits the operator's next target, ready to morph back into UGV if its next goal is reachable by ground mobility. 

Three waypoints were given to the robot to follow, where the first and last waypoints can be reached by ground mobility while the second waypoint crosses the wall for the robot to fly over. Throughout the experiments, the operator only provided the robot with the waypoints while the robot autonomously performed the control, waypoint tracking, and transformation. We attached the robot to a safety tether to prevent a crash in case of emergency or controller failure in the air. We also attached ethernet and data cables to the robot for data logging which we can use to plot the tracking performance, alongside the localization and odometry visualization by the stereo depth camera.

The results of the experiment can be seen in Fig.~\ref{fig:composite}, \ref{fig:odometry}, and \ref{fig:odometry_plot}. Figure~\ref{fig:composite} is a composite image showing the robot's path as it flew over the wall in UAV mode, landed on the other side of the wall, transformed back into the UGV mode, then drove to the last waypoint. Figure~\ref{fig:odometry} shows the stereo depth camera's visualization of the detected environment, the odometry using SLAM, path planning using the A$^*$ algorithm, and the path taken by the robot. Figure~\ref{fig:odometry_plot} shows the SLAM odometry using the stereo depth camera, desired goal, aerial waypoints, and Ardupilot's EKF pose estimates while in the air. 
%
As shown in Fig.~\ref{fig:odometry}, the robot has successfully reached the desired waypoints using only the odometry from the onboard sensors. The ground trajectory tracking performance worked well, as shown in Fig.~\ref{fig:odometry_plot}, where the robot can reach the target position using simple forward and turning speed commands. On the other hand, aerial trajectory tracking has some oscillations and overshoots, which are pronounced in the altitude and heading measurements. Our tests showed that there is approximately a second delay between when the command is sent and the robot's actual movement.
This delay is likely caused by the limitation of either the stereo camera or the onboard computer. This delay could cause a stability issue for longer flights, therefore we kept the flight time short which works well for the simple trajectory that we used in the experiment.

\section{Conclusions and Future Work}


In this paper, we presented our multi-modal morphing robot and the implementation of SLAM, path planning, and trajectory tracking using only the onboard computer and sensors in an indoor environment. The robot utilized its onboard stereo depth camera to perform SLAM and autonomously navigated a complex indoor environment, and transformed between UGV and UAS modes to drive and fly over obstacles. Experimental results show that the robot can estimate its pose using vision-based SLAM, perform path planning, track the waypoints generated by the path planner, and navigate to the target position using both ground and aerial modes. There are some issues that need to be addressed in future work, such as the pose estimation delays that can cause instability during flight. Once these issues have been resolved, we can then proceed to conduct outdoor experiments to show that the robot can perform well in an outdoor unstructured environment.








\balance{}
 
\bibliographystyle{IEEEtran}
\bibliography{references}
\end{document}